\documentclass[letterpaper, 10 pt, conference]{ieeeconf}  % Comment this line out if you need a4paper

\IEEEoverridecommandlockouts                              % This command is only needed if 
\usepackage[colorinlistoftodos]{todonotes}

\usepackage{caption}
\usepackage{subcaption}
\usepackage{amsmath}
\usepackage{amsfonts}
\usepackage{flushend}
\usepackage[hidelinks]{hyperref}
\newcommand{\overbar}[1]{\mkern 1.5mu\overline{\mkern-2.5mu#1\mkern-0.5mu}\mkern 1.5mu}

\title{\LARGE \bf
ViT-A*: Legged Robot Path Planning using Vision Transformer A*
}

\author{Jianwei Liu$^{*}$, Shirui Lyu$^{*}$, Denis Hadjivelichkov, Valerio Modugno, and Dimitrios Kanoulas% <-this % stops a space
\thanks{The authors are with the Department of Computer Science, University College London, Gower Street, WC1E 6BT, London, UK. {\tt\small \{jianwei.liu.21, shirui.lyu.19, dennis.hadjivelichkov, v.modugno, d.kanoulas\}@ucl.ac.uk}}% <-this % stops a space
\thanks{*equal contribution}% <-this % stops a space
\thanks{This work was supported by the UKRI Future Leaders Fellowship [MR/V025333/1] (RoboHike) and the CDT for Foundational Artificial Intelligence [EP/S021566/1].  For the purpose of Open Access, the author has applied a CC BY public copyright license to any Author Accepted Manuscript version arising from this submission.}}

\begin{document}

\maketitle
\thispagestyle{empty}
\pagestyle{empty}

%%%%%%%%%%%%%%%%%%%%%%%%%%%%%%%%%%%%%%%%%%%%%%%%%%%%%%%%%%%%%%%%%%%%%%%%%%%%%%%%
\begin{abstract}
Legged robots, particularly quadrupeds, offer promising navigation capabilities, especially in scenarios requiring traversal over diverse terrains and obstacle avoidance. This paper addresses the challenge of enabling legged robots to navigate complex environments effectively through the integration of data-driven path-planning methods. We propose an approach that utilizes differentiable planners, allowing the learning of end-to-end global plans via a neural network for commanding quadruped robots. The approach leverages 2D maps and obstacle specifications as inputs to generate a global path. To enhance the functionality of the developed neural network-based path planner, we use Vision Transformers (ViT) for map pre-processing, to enable the effective handling of larger maps. Experimental evaluations on two real robotic quadrupeds (Boston Dynamics Spot and Unitree Go1) demonstrate the effectiveness and versatility of the proposed approach in generating reliable path plans.
\end{abstract}

%%%%%%%%%%%%%%%%%%%%%%%%%%%%%%%%%%%%%%%%%%%%%%%%%%%%%%%%%%%%%%%%%%%%%%%%%%%%%%%%
\section{INTRODUCTION}
Legged robots, and especially quadrupeds, have seen tremendous progress over the past few years allowing them to carry out a wide range of tasks, ranging from package delivery~\cite{ji2023dribblebot}, agricultural production~\cite{Dandelionrobot}, and search-rescue missions~\cite{liu2007legged}. Path planning plays a crucial role in enabling legged robots to navigate autonomously and effectively in various complex environments. Several studies, e.g.,~\cite{Lau2015, Zhang_2022}, were dedicated to the development of efficient path planning algorithms for quadrupedal robots, aiming to ensure their safe navigation while avoiding collisions with obstacles. Many of these works have utilized traditional methods such as Rapidly-exploring Random Trees (RRT) and $A^{*}$-based methods. Despite all the great efforts, achieving efficient and reliable path plans for mobile robots continues to present an ongoing challenge when using such traditional approaches. For example, established planning methods often struggle to effectively handle the complexities and uncertainties associated with real-time sensor inputs~\cite{choudhury2018data}. 

\begin{figure}[t!]
    \centering
       \subfloat[Boston Dynamics Spot]{%
          \includegraphics[height=4.cm]{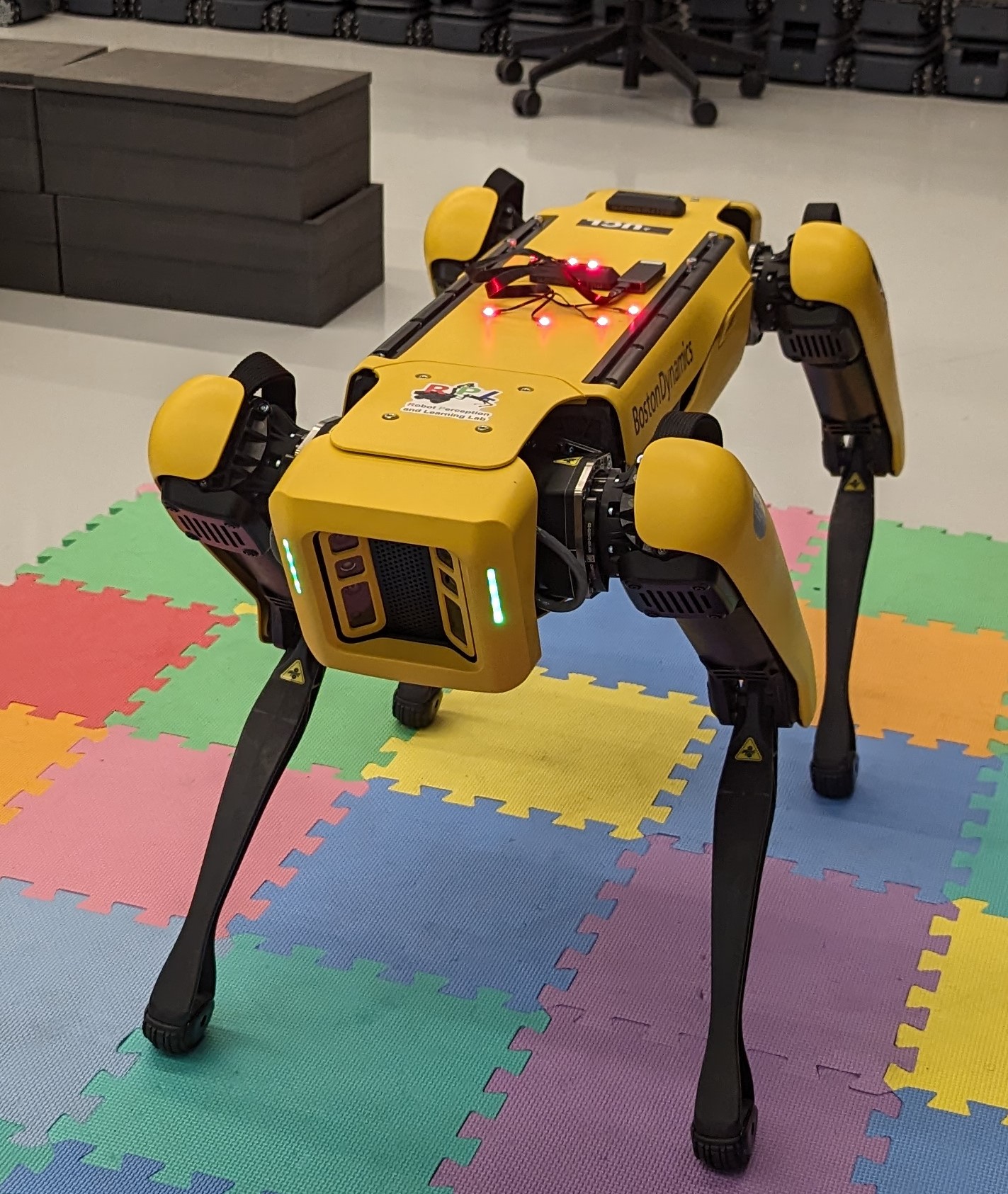}%
          \label{subfig:spot_with_PS_markers}%
       } 
       \quad
       \subfloat[Unitree Go1]{%
          \includegraphics[height=4.cm]{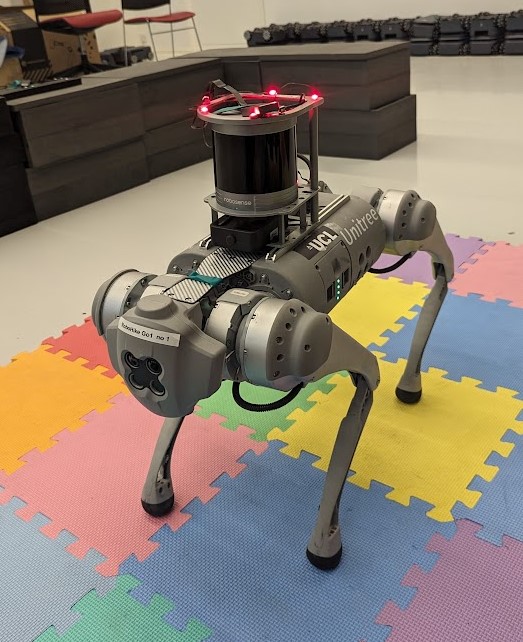}%
          \label{subfig:go1_with_PS_markers}%
       }
    \caption{The two robots (left: Boston Dynamics Spot, right: Unitree Go1) used for validating the proposed method.}
    \label{fig:robots_with_PS_markers}
\end{figure}

In contrast, the emergence of planners that integrate data-driven methods, fostered by the advancements in Deep Learning (DL), offers a promising avenue for addressing some of these challenges, e.g., empowering robots to learn and adapt from real-world data~\cite{yonetani2021path, Pogancic2020}. This ability to gather knowledge from real-world experiences equips robots with the capacity to make more informed decisions in diverse and challenging situations.
 
In this paper, we present an approach that builds upon recent advancements in differentiable planners~\cite{Pogancic2020}, enabling the learning of end-to-end mapping. Specifically, we focus on generating global paths for quadrupedal robots, by feeding 2D maps with obstacles into a deep neural network for $A^*$-based learning. Moreover, we enhance the functionality of our neural network-based planner by using a map pre-processing step with Visual Transformers (ViT)~\cite{dosovitskiy2020image}. By introducing such an encoding, our method can leverage the strengths of transformers, particularly in capturing long-range dependencies and learning complex relationships in the input map images, while enabling the handling of larger maps efficiently. In the remainder of this paper, we refer to the proposed method as \textit{ViT-$A^*$} Path Planner. The contributions of this work can be summarized as follows. We introduce:
\begin{itemize}
    \item a ViT-based Neural $A^*$ Path Planner (ViT-$A^*$) that operates efficiently on maps of any dimension;
    \item a control stack to ensure the successful application of the proposed method on real quadruped robots in numerous application scenarios.
\end{itemize} 

The subsequent sections are structured as follows. Sec.~\ref{sec:rw} provides an overview of the relevant literature, discussing previous works in the field, while Sec.~\ref{sec:method} outlines the proposed method in detail. Sec.~\ref{sec:exp} presents the experimental setup, including both simulation and real robot experiments. Finally, in Sec.~\ref{sec:conclusion}, we present the conclusions and discuss future developments of our research.
\section{RELATED WORK}\label{sec:rw}
\subsection{Classical Path Planning}
There are two main approaches to classical path planning algorithms: search-based and sampling-based methods. Search-based path planning provides mathematical guarantees of converging to a solution if it exists.  $A^*$ and its modifications have since found extensive use in robot navigation due to their simplicity in implementation and effectiveness in finding valid paths. For instance, in~\cite{Huang2022} the authors introduced an extension of  $A^*$ to drive a mobile platform to sanitize rooms. In~\cite{raghavan2020agile, Kusuma2019} $A^*$ algorithms were used to find collision-free paths for a legged or legged-wheeled~\cite{raghavan2019variable, sushrutha2021reconfigurable} robots to achieve autonomous navigation. Extensions to these include footstep perception~\cite{kanoulas2017vision, Stumpf2022, kanoulas2019curved} and planning~\cite{kanoulas2018footstep}, or even navigation among movable obstacles~\cite{ellis2022navigation, ellis2023navigation, linghong2023local}. Traditional methods heavily rely on a fixed heuristic function, such as the Euclidean distance, which lacks adaptability to varying robot configurations and environments. In our work, we propose a novel approach where we learn a heuristic based on the visual appearance of the application scenarios allowing the robot to make more informed decisions and thus reducing the overall search area and planning time.

Sampling-based planners efficiently create paths in high-dimensional space by sampling points in the state space. They can effectively work with continuous spaces. The literature in this context is vast, especially for applications in legged robotics. Some notable contributions in the field include~\cite{Lau2015}, where an extension of an RRT-based algorithm is used for controlling a quadruped robot during the DARPA Challenge in 2015. More recently~\cite{Zhang_2022} introduced a novel sampling-based planner that shortens the computational time to find a new path in quadrupedal robots. While these approaches demonstrate satisfactory performance and probabilistic convergence, their limitations lie in the inability to incorporate image-based information directly into the planning process. As a result, their application is restricted in scenarios where planning based on visual data is not essential.

\begin{figure*}[t!]
    \centering
    \includegraphics[width=1\textwidth]{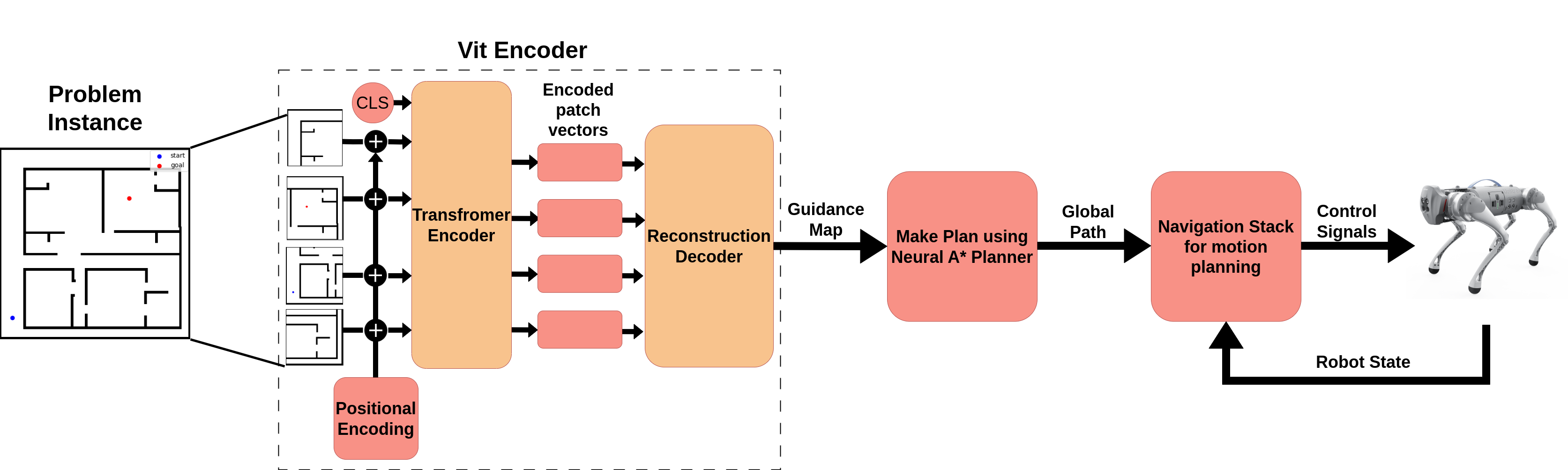}
    \caption{Overall system, tested on the real robots. The 2D map is decomposed into patches and then fed to the ViT module. After the encoding-decoding process, the resulting Guidance Map is given to the $A^*$ and it is used to find a global path. Finally, the global path is executed by the navigation stack, which controls the real robot to ensure small tracking errors.}
    \label{fig:nerual_a-star_arch}
\end{figure*}

\subsection{Data-Driven Path Planning}
In contrast to the classical path-planning methods, state-of-the-art research in the field has shifted towards more practical solutions, which involve incorporating machine learning techniques. Data-driven methods have emerged as robust solutions to address these challenges by directly learning the behavior of pathfinding. These methods employ approaches such as expert demonstration~\cite{pfeiffer2017perception} or imitation learning~\cite{choudhury2018data} to learn how to plan paths. Recent works directly address the issue of lack of semantically labeled maps in classical search-based methods by using data-driven approaches directly on raw image~\cite{ichter2019robot, choudhury2018data, lee2018gated}. Specifically, Yonetani et al.~\cite{yonetani2021path} introduced Neural $A^*$ -- a differentiable variant of the canonical $A^*$, coupled with a neural network trained end-to-end. The method works by encoding natural image inputs into guidance maps and searching for path-planning solutions on them, resulting in significant performance improvements over previous approaches both in terms of efficiency and optimality. Our work expands upon this paper.

\subsection{Vision Transformers}
While methods such as the Neural $A^*$ ~\cite{yonetani2021path} have shown great promise in terms of performance improvements, they face limitations in processing larger maps due to the use of Convolutional Neural Networks (CNNs), where dealing with maps with increasing size could lead to a reduction in performance. This has posed some constraints in terms of processing larger maps. Transformers have emerged as a promising alternative, exhibiting significant performance improvements in various computer vision tasks~\cite{codetr2022, wang2023one, Jinjing2023} and robot vision tasks~\cite{ha2022semabs, hao2020prevalent, Hadjivelichkov2022affcorrs}. Transformers have the ability to capture long-range dependencies in images, thanks to their self-attention layers that enable them to attend to any part of the image regardless of the distance from the current location~\cite{dosovitskiy2020image, Raghu2021DoVT}. This is in contrast to CNNs, which are confined to focus on local image patches. Moreover, due to their stacked layers, transformers can learn more complex relationships between different parts of the images while assuming fewer inductive biases~\cite{Raghu2021DoVT}. In this work, we exploit the capability of the transformers to learn long-range dependencies to enhance the Neural A* performances with larger maps.
\section{ViT-BASED NEURAL $A^*$ PATH PLANNER}\label{sec:method}

\subsection{Neural $A^*$ Planner}\label{Sec:NAStart}
This work expands upon the Neural $A^*$ Path Planner, introduced in~\cite{yonetani2021path}. Our method aims to provide global path plans as depicted in Fig.~\ref{fig:nerual_a-star_arch}. In our approach, we introduce a ViT network, instead of the original CNN-based encoder-decoder structure, to process 2D maps of the environment. Unlike classification tasks using transformers that output a fixed-sized vector, our design allows a path planner to operate on variable-sized map inputs. To achieve this, we incorporate a decoder architecture that converts the embedded vectors for individual image patches back to the required guidance map. By utilizing the attention mechanism, our planner can effectively focus on key features in the planner task, such as obstacles, as well as start and goal positions, while exploiting the differentiable $A^*$'s ability to learn the decoding efficiently.

Neural $A^*$ is a path planning algorithm that combines the convergence guarantees of $A^*$ with the flexibility that characterizes a neural network to learn how to exploit visual cues in order to find near-optimal paths. In our setting, a $i$-th path planning problem is defined as
\begin{equation}
  {Q^{i} = (X^i, v_s^i,v_g^i, \overbar{P}^{i})},
\end{equation}
where $X^i$ represents a 2D map of the current scenario, $ v_s^i$ and $v_g^i$ are respectively the start and goal position, and $\overbar{P}^{i}$ is a ground truth binary map representing the desired path. The Neural $A^*$ path planner is composed of distinct sequential steps. Firstly, the 2D map $X^i$ which has dimensions of $H \times W \times C$, where H and M are the dimensions of the map and C indicates the number of color channels ($C = 3$ for RGB maps and $C = 1$ for binary occupancy maps) is fed to a CNN-based encoder. The encoder learns to map the raw image input to a guidance map defined as
\begin{equation*}
  f: \mathbb{R}^{H \times W \times C} \rightarrow \mathbb{R} ^ {H \times W}.
\end{equation*}
The guidance map represents the cost of traveling to adjacent nodes in the map, which is equivalent to the sum of the heuristic cost and the grid travel cost in regular $A^*$ algorithms. Finally, a minimal cost path is found, following the guidance map and using the traditional $A^*$ algorithm to explore the search space and find a valid path.

The differentiability of the path-finding process in the neural $A^*$ path planner plays a crucial role in enabling the training of the CNN-based encoder pre-processor. This allows the neural network to learn and capture the essential features and patterns required for efficient path planning. The differentiability is achieved through a full matrix reformulation of the $A^*$ algorithm, enabling the computation of gradients accounting for every search step during the backpropagation stage. In this paper, we focus solely on the node selection step for the original $A^*$ for the sake of simplicity (for Neural $A^*$ the cost terms in the node selections step are replaced with the guidance map). Therefore, given the regular $A^*$ node selection rule:
\begin{equation}
  v^* = {argmin}_{v \in \mathcal{O}} \left(g(v) + h(v)\right),
\end{equation}
where $\mathcal{O}$ represents the list of candidate nodes (assuming a 2D map represented a graph where each pixel is a node), $g(v)$ refers to the accumulated total cost on the optimal path up to the node $v$, and $h(v)$ is the heuristic function that provides an estimation from the candidate node to the goal, the node selection step of the $A^*$ path planner can be redefined in a matrix form as:
\begin{equation} \label{eq:N_a_star_node_select}
  \mathcal{V}^\ast = \mathcal{I}_{max} \left( \frac{exp\left(-(G+ H) / \tau\right) \odot O}{\langle exp\left(-(G+ H) / \tau\right), O \rangle} \right),
\end{equation}
where the functions $G$ and $H$ are the matrix formulation of the functions $g(v)$ and $h(v)$ respectively. The one-hot-encoding scheme is used in the following steps, which acts as a matrix mask such that only the selected node has the value one and zero otherwise. The one-hot-encoding for the next optimal node is denoted as $\mathcal{V}^\ast$. The parameter $\tau$ is determined empirically, and the symbol $A \odot B$ indicates an element-wise product between matrices $A$ and $B$. During the forward pass, $\mathcal{I}_{max}$ is determined using the $argmax$ function, while during back-propagation, it is treated as an identity.

The loss function is computed as the average $L_1$ loss between the selected nodes  from the $A^*$ denoted as $P$ (which represents a global path), and the ground-truth path map $\overbar{P}^{i}$, which is given as input:
\begin{equation}
  \mathcal{L} =\lVert P - \overbar{P}^{i} \rVert_1/\mathcal{V}
  \label{eq:loss}
\end{equation}
This loss function serves to supervise the guidance map driven nodes selection by penalizing two types of errors: the absence of nodes that should have been included in $P$ to correctly reconstruct $\overbar{P}^{i}$, and the presence of an excessive number of nodes in $P$ that do not belong to $\overbar{P}^{i}$. 

\subsection{Vision Transformer for Guidance Map Encoding}\label{Sec:ViTAStar}
To achieve a sophisticated and attention-based encoding for raw image inputs, we have introduced a Vision Transformer (ViT)~\cite{dosovitskiy2020image} to extend the capability of the proposed method. The purpose of the model is to encode an input image into a guidance map by taking into account the visual cues. Detailed schematics of the ViT module can be found in Fig.~\ref{fig:nerual_a-star_arch}.

\begin{figure*}[t!]
\centering
   \subfloat[office01\\($280\times280$)]{%
    \includegraphics[width=0.15\textwidth]{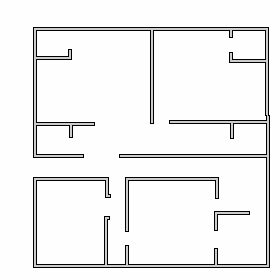}
      \label{subfig:office01}%
      } 
       \hfill
   \subfloat[room02\\($360\times360$)]{%
     \includegraphics[width=0.15\textwidth]{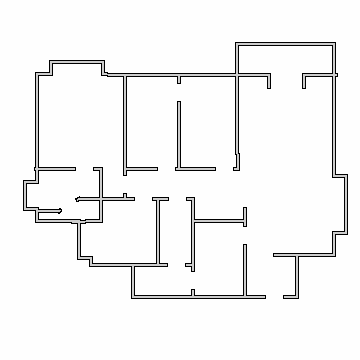}
      \label{subfig:map_maze}%
      } 
       \hfill
   \subfloat[office02\\($600\times600$)]{%
     \includegraphics[width=0.15\textwidth]{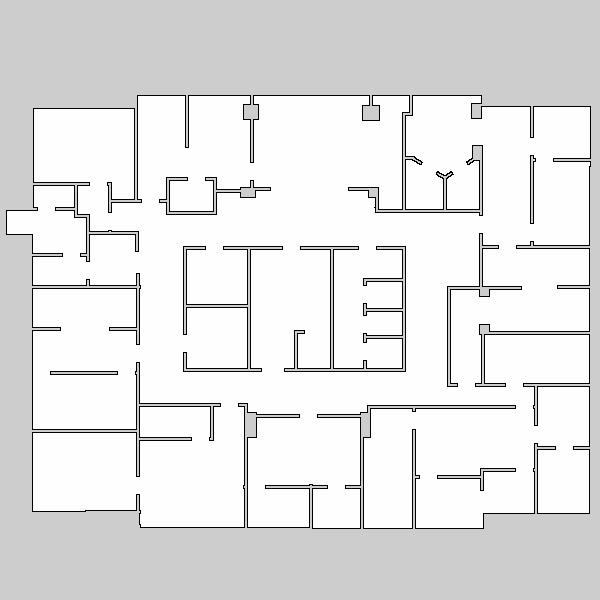}
      \label{subfig:map_office02}%
      } 
       \hfill
   \subfloat[maze\\($600\times600$)]{%
      \includegraphics[width=0.15\textwidth]{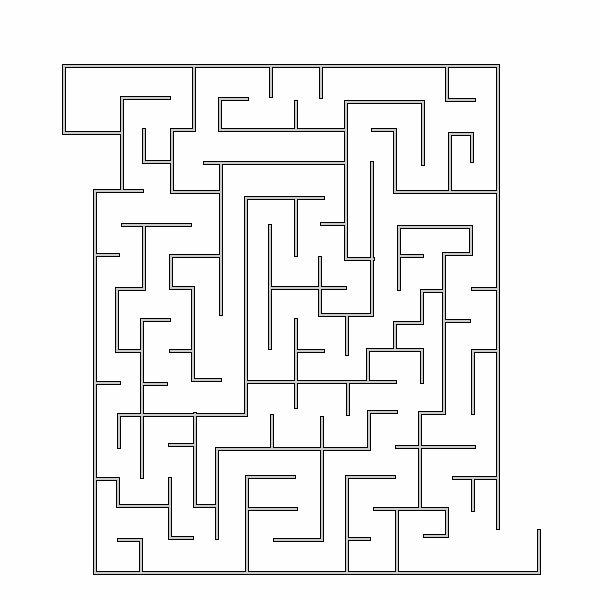}
      \label{subfig:map_maze}%
      } 
       \hfill
    \subfloat[mall\\($760\times760$)]{%
      \includegraphics[width=0.15\textwidth]{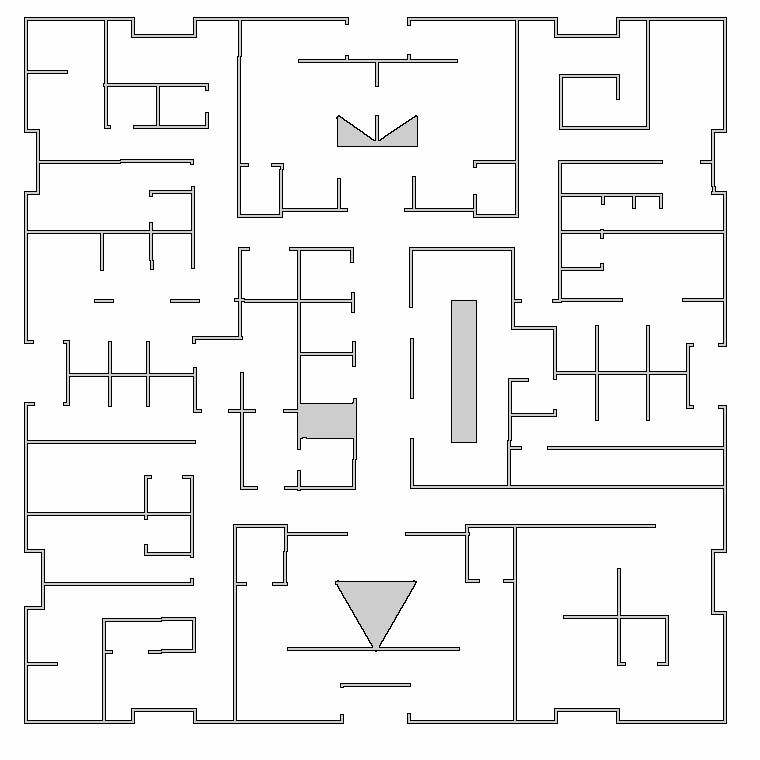}
      \label{subfig:map_mall}%
      }
    \caption{Maps used for comparing different planning methods with their sizes.}
    \label{fig:maps_bench}
\end{figure*}

The input map is initially represented as a tensor
\begin{equation*}
  X^i \in \mathbb{R}^{H \times W \times C},
\end{equation*}
where $H$, $W$, and $C$ have already been defined in Sec~\ref{Sec:NAStart}. Since the ViT module expects a sequential input, we reshape the map matrix into a flattened sequence of 2D patch vectors denoted as $x_s$, with the shape of $x_s \in R^{N \times (S^2 \cdot C)}$. Here, $S$ represents a hyper-parameter indicating the patch dimension, and $N = HW / S^2$ is the resulting number of patches from the map input. 

To ensure that the input map size is compatible with the required patches, we have introduced padding. This is necessary when the dimensions of the input map, $H$ and $W$, are not an integer multiple of patch size $S$, and thus may not be segmented into the required patches. 

In the sequence of patches, we incorporate a positional embedding, following the approach used in the ViT models~\cite{dosovitskiy2020image}. This positional embedding serves to indicate the positional relationship between the patches, mimicking the spatial information presented in the original raw image.

To address the challenge of variable size inputs, we follow the idea proposed in the work presented in~\cite{wang2022dm} where a positional upper bound representing the maximum number of possible patches is introduced. This ensures that the model can handle inputs of varying sizes without sacrificing training efficiency. 
Subsequently, each vector in the sequence is subjected to encoding, leading to the generation of an embedded vector projected into the hidden dimension. This encoding process effectively converts the input patches into a latent representation that captures their significant features.

Finally, the embedded vector sequence is decoded into vectors of size $S^2$ using the reconstruction decoder. The purpose of this decoder is to reconstruct the guidance map, which is required to have the same dimensions as the input map. The guidance map is a crucial component as it provides essential information for the planning cost. Each individual entry in the guidance map represents a corresponding guidance cost for planning, facilitating the decision-making process based on the encoded visual cues captured by the model.

\begin{figure}[b!]
    \centering
    \subfloat[Problem Instance]{%
    \includegraphics[width=0.2\textwidth]{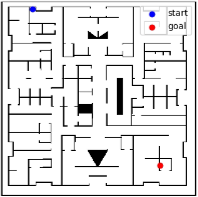}
    \label{subfig:planning_comparison_problem_instance}
    } 
    \subfloat[Regular $A^*$]{%
    \includegraphics[width=0.2\textwidth]{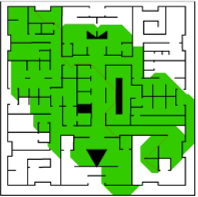}
    \label{subfig:planning_comparison_regular_a-star}
    } 

    \subfloat[Neural-$A^*$]{%
    \includegraphics[width=0.2\textwidth]{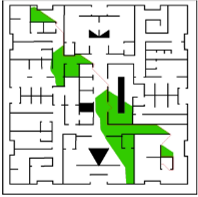}
      \label{subfig:planning_comparison_NN_a-star}
      } 
    \subfloat[ViT-$A^*$]{%
    \includegraphics[width=0.2\textwidth]{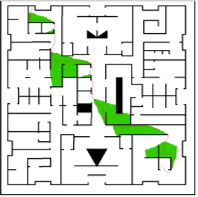}
    \label{subfig:planning_comparison_ViT}
    } 
    \caption{Visualization of planning results by Regular A*, Neural A*, and ViT A*. In greem, the search area.}
    \label{fig:planning_comparison}
\end{figure}

\section{EXPERIMENTAL RESULTS}\label{sec:exp}
In this section, we test our method. Firstly, in Sec.~\ref{Sec:Benchmark}, we run a path planning comparison between standard $A^*$, Neural Network-based $A^*$ and ViT-based $A^*$. In Sec.~\ref{Sec:realexp}, we test our method on two real robots, Boston Dynamics Spot and Unitree Go1.

\subsection{Benchmarking Comparison}\label{Sec:Benchmark}
\begin{table}[ht!]
    \centering
    \begin{tabular}{ c c c c} 
     \hline
     maps & ViT-$A^*$ & N-$A^*$ &  $A^*$ \\ [0.5ex] 
     \hline\hline
     (a) & 5.68             & \textbf{4.70} & 6.03 \\ 
     (b) & 17.31            & \textbf{14.73} & 17.51 \\  
     (c) & \textbf{4.81}    & 5.17 & 15.59 \\
     (d) & \textbf{69.97}   & 75.20 & 84.84 \\
     (e) & \textbf{12.73}   & 16.57 & 36.24 \\[1ex] 
     \hline
    \end{tabular}
    \caption{\textbf{Planning time}: Average run-time (in sec) required to solve a single planning problem on maps from Fig.~\ref{fig:maps_bench}.}
 \label{tab:sim_run_time_comparision}
\end{table}

Here, we compare our proposed ViT-based $A^*$ path planner (ViT-$A^*$) as described in Sec.~\ref{Sec:ViTAStar}, against two baselines: the Neural Network-based $A^*$ (N-$A^*$)~\cite{yonetani2021path} and a classic $A^*$ planner, as described in Sec.~\ref{Sec:NAStart}. The evaluation is based on 2D maps coming from the MRPB benchmark dataset~\cite{Wen2021}. As we enabled our planner to work on variable map dimensions without specific training, the maps selected have different sizes that range from $280 \times 280$ to $760 \times 760$ pixels, and are all depicting realistic scenarios, such as offices or rooms (Fig.~\ref{fig:maps_bench}), which helps to bridge the reality gap when deploying the proposed method on real quadrupedal robots. 

\begin{figure*}[t!]
\centering
   \subfloat{%
    \includegraphics[width=0.243\textwidth]{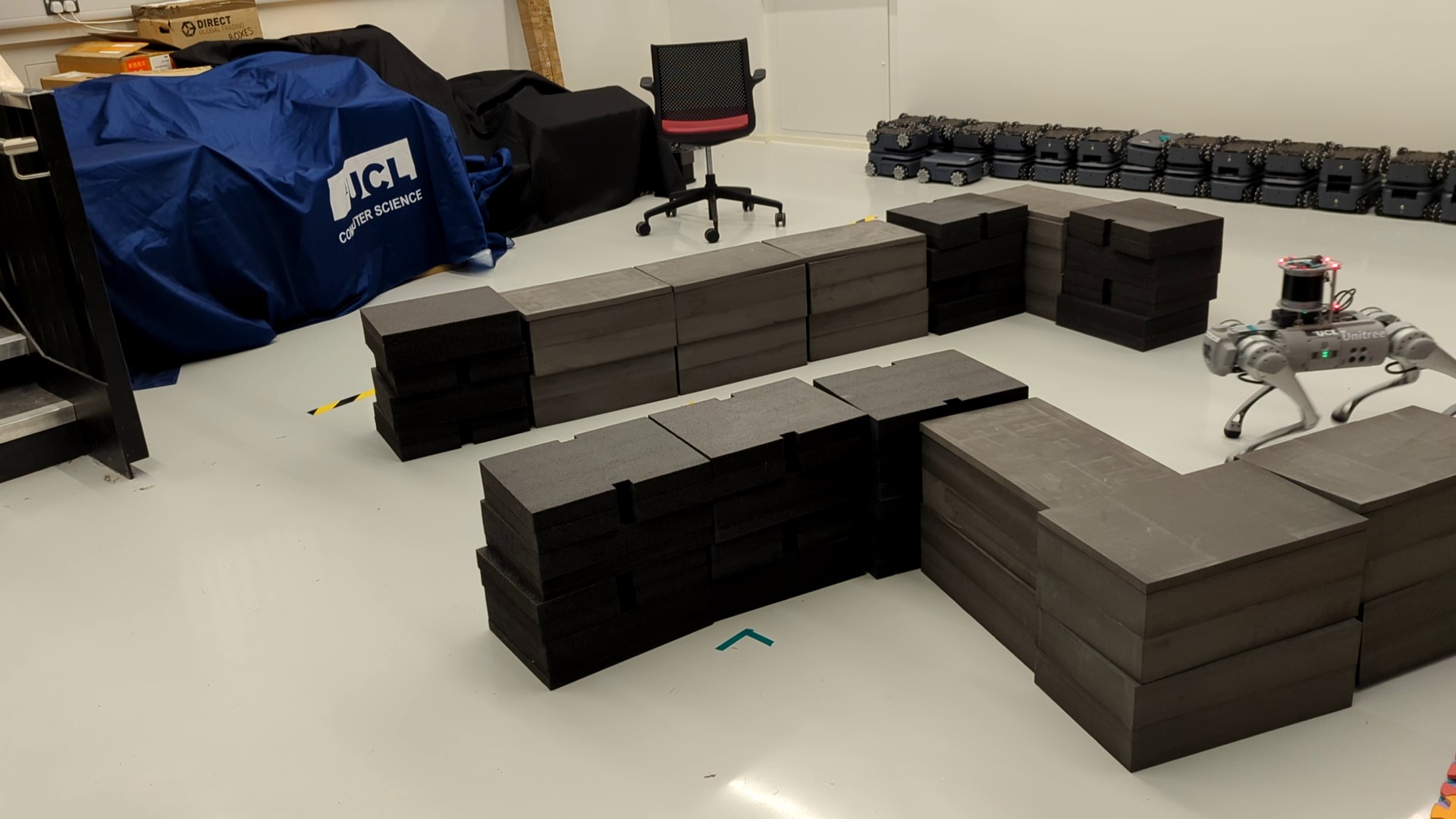}
      \label{subfig:Go1_nav_around_obstacle_1}% 0.243
      } 
   \subfloat{%
     \includegraphics[width=0.243\textwidth]{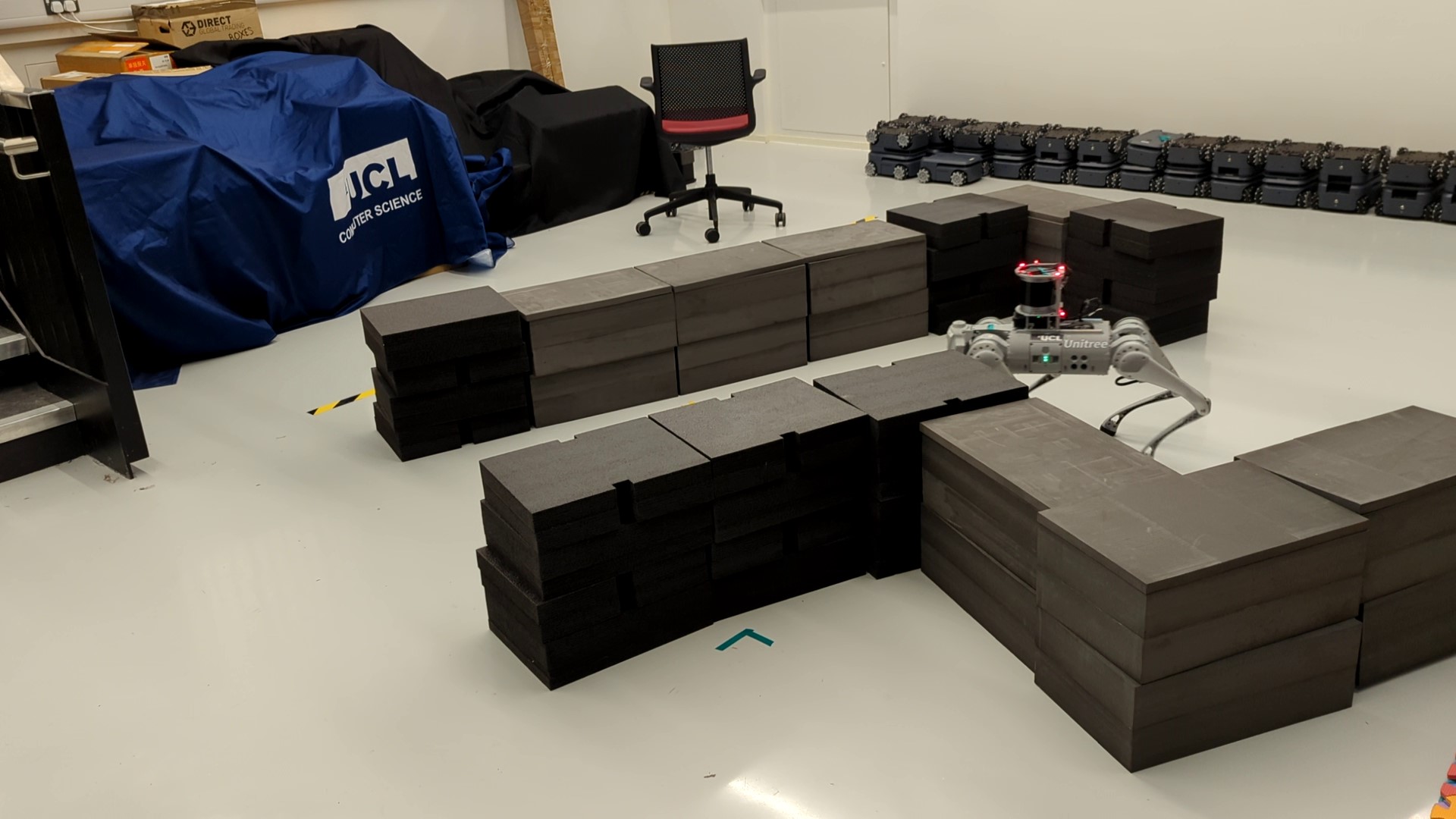}
      \label{subfig:Go1_nav_around_obstacle_2}%
      } 
   \subfloat{%
     \includegraphics[width=0.243\textwidth]{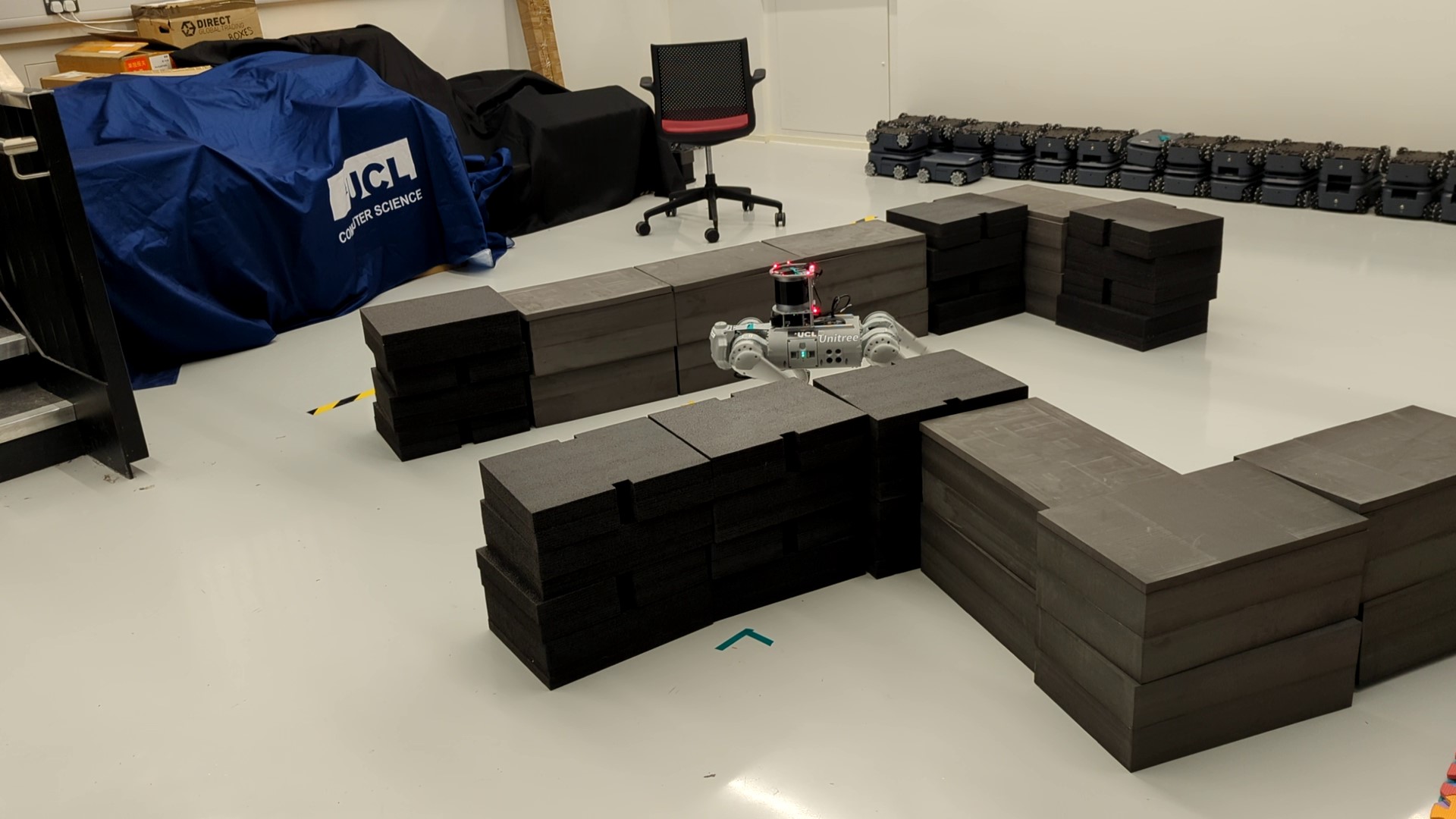}
      \label{subfig:Go1_nav_around_obstacle_3}%
      } 
   \subfloat{%
      \includegraphics[width=0.243\textwidth]{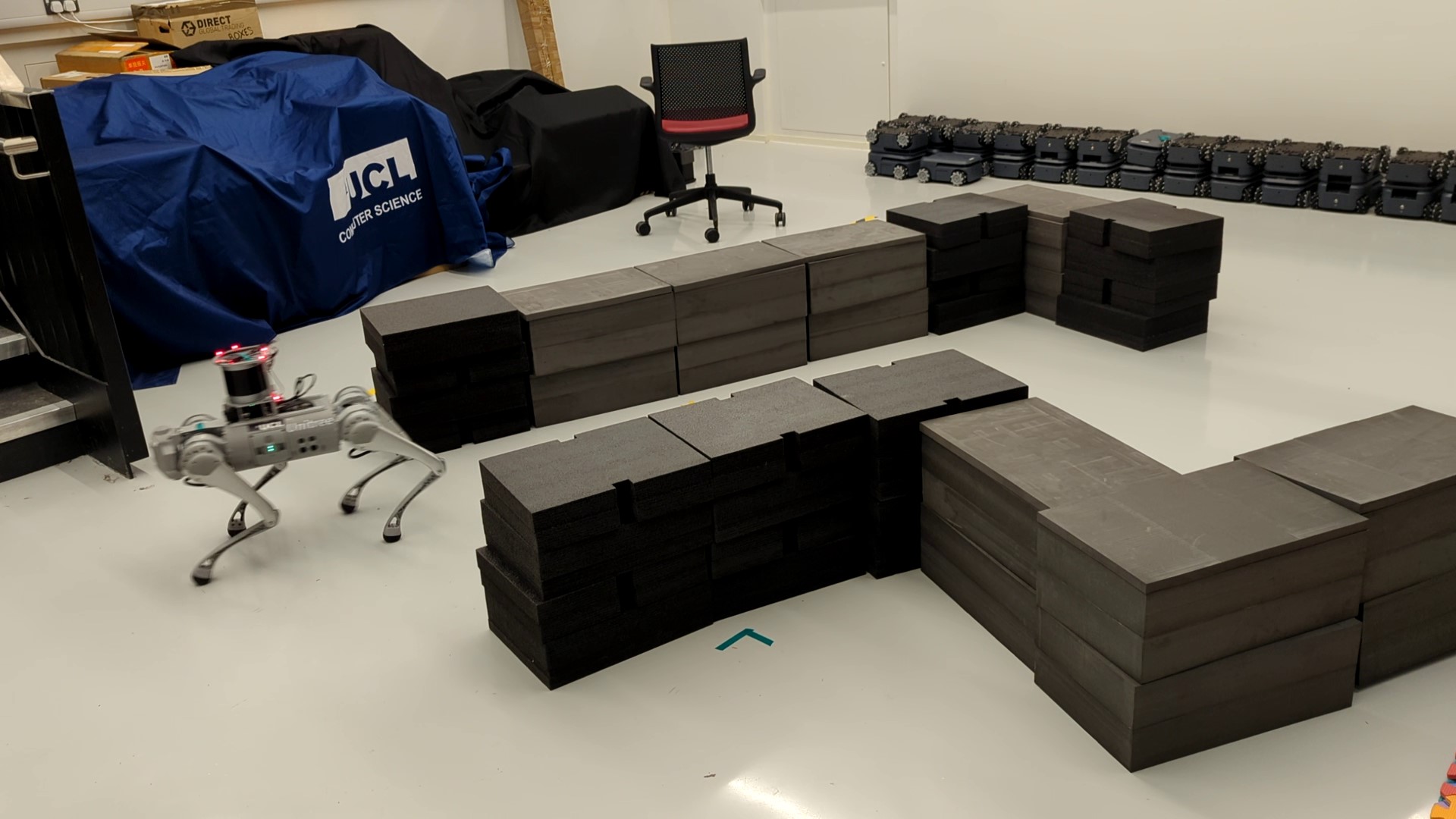}
      \label{subfig:Go1_nav_around_obstacle_4}%
      }
      \vspace{0.1cm}
     \subfloat{%
      \includegraphics[width=0.243\textwidth]{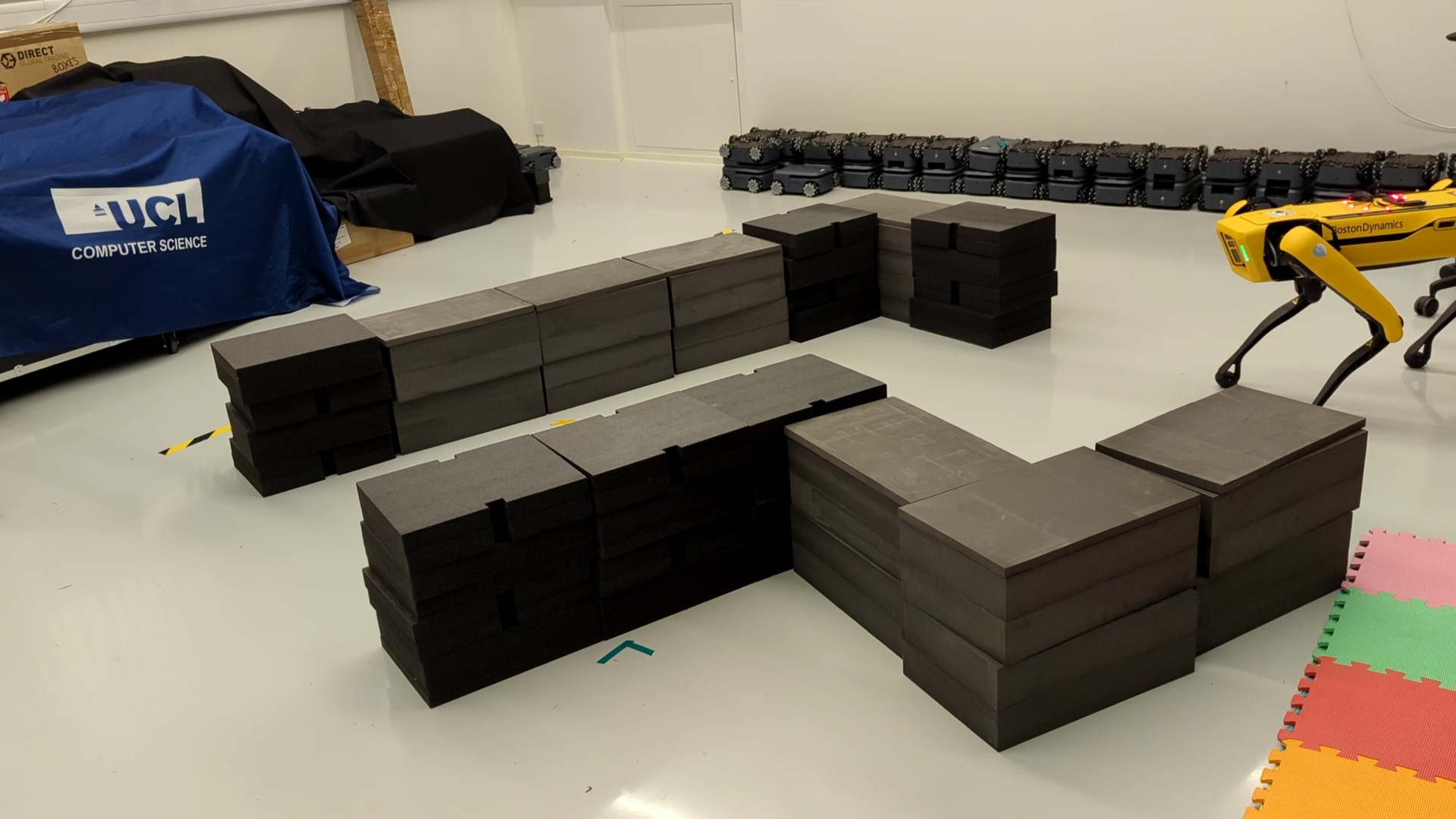}
      \label{subfig:spot_nav_around_obstacle_1}%
      }      
    \subfloat{%
      \includegraphics[width=0.243\textwidth]{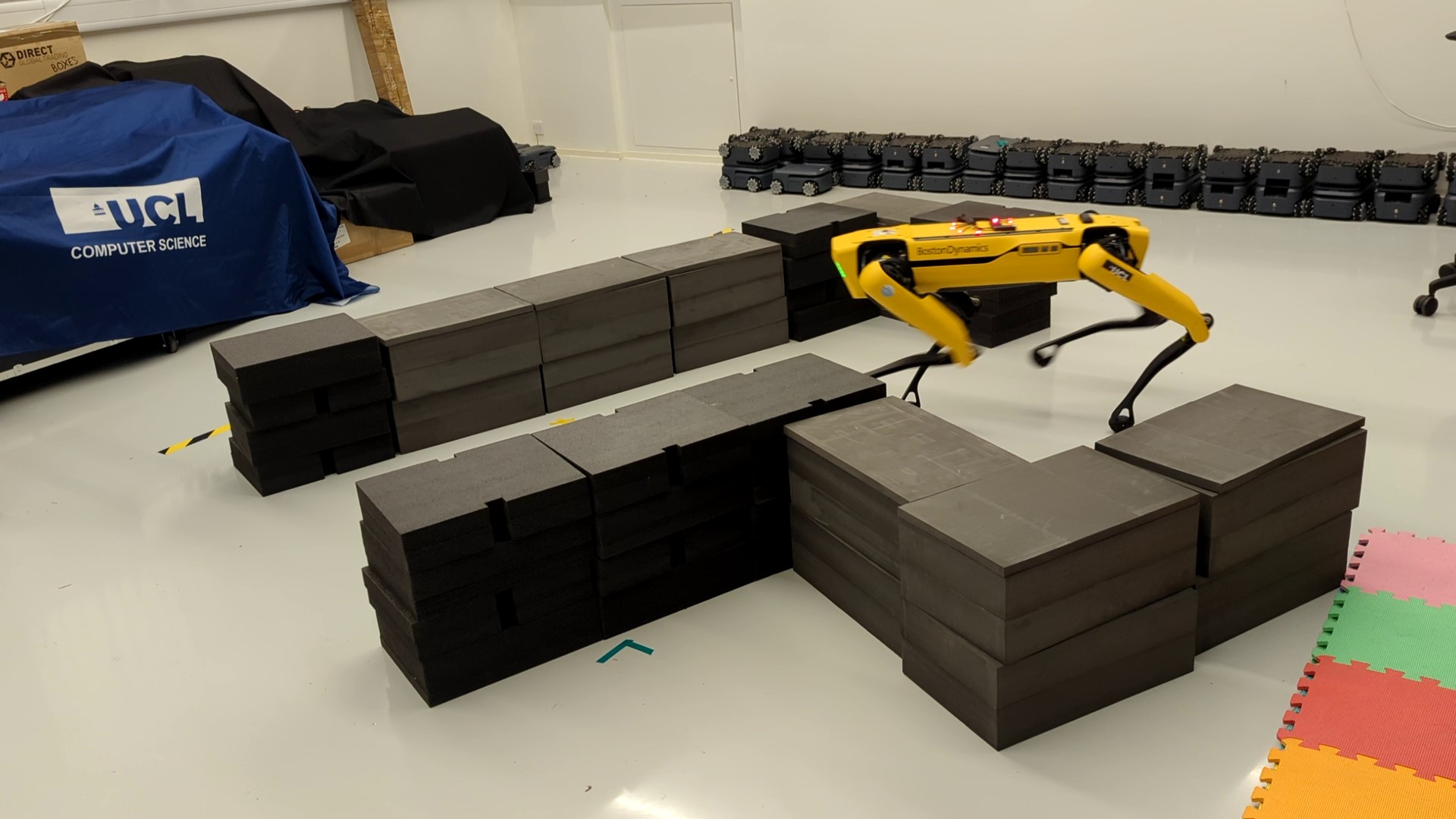}
      \label{subfig:spot_nav_around_obstacle_2}%
      }      
    \subfloat{%
      \includegraphics[width=0.243\textwidth]{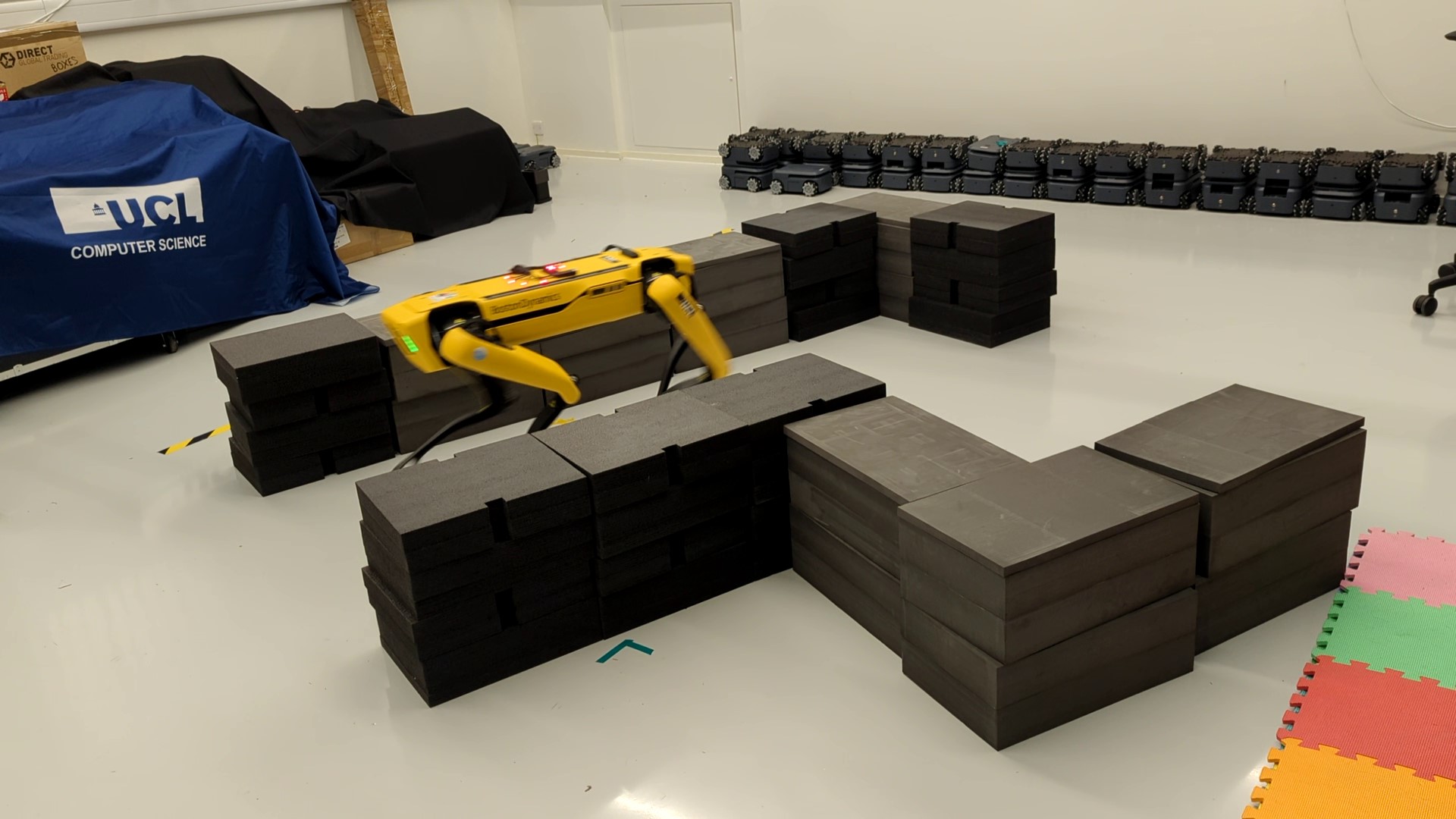}
      \label{subfig:spot_nav_around_obstacle_3}%
      }      
    \subfloat{%
      \includegraphics[width=0.243\textwidth]{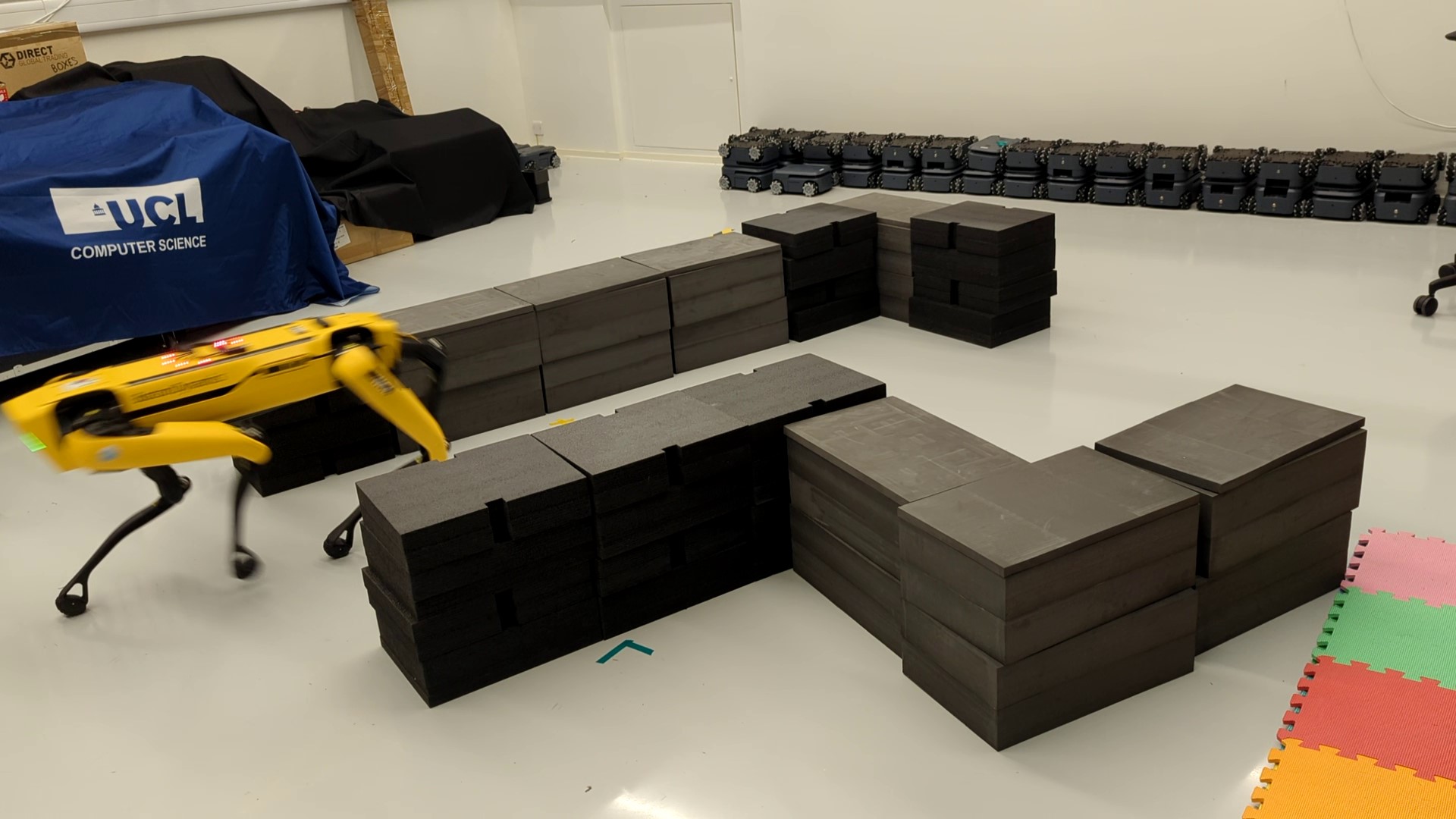}
      \label{subfig:spot_nav_around_obstacle_4}%
      }

    \caption{From left to right we show a sequence of the Unitree Go1 (top) and Boston Dynamics Spot (bottom) robots navigating around obstacles using the ViT-$A^*$ driven navigation stack.}
    \label{fig:Go1_nav_around_obstacle}
\end{figure*}

To guarantee an effective comparison, we defined a precise generation protocol for testing cases. First, to test the planners' generalization capability (especially for maps with different sizes) we generate random samples of start and goal positions. The randomly generated start and goal must not intersect with obstacles in order to define a valid test case. This constraint, together with the fact that there are no closed regions in the map we used, ensures the completeness of the plan so it is ensured that a solution path exists. Furthermore, to avoid trivial planning tasks that are too short in length, we force the generation process to separate the start and goal positions by a threshold value. Hence, every planning task exceeds a certain length in the experiments. 

In the experiments, we applied the loss function defined in 
Eq.~\ref{eq:loss} to train both models. To train the models, the RMSprop optimizer is selected and for both models, a learning rate of 0.001 is used. The CNN model and the ViT model are trained on the same dataset for 300 epochs or until converge criteria are met.

Each planner is evaluated by considering the \textit{planning time} metric. The planning time is a key feature in evaluating the quality of a planner since it is a measure of the algorithm's efficiency. 

The results shown in Table~\ref{tab:sim_run_time_comparision} are obtained by running each planner on the maps shown in Fig.~\ref{fig:maps_bench}. For each map, we compute the average planning time by repeating the path planning task 25 times with different start and goal positions.
from Table~\ref{tab:sim_run_time_comparision} it is clear how our approach outperforms N-$A^*$ and $A^*$ especially for maps with larger size. Moreover, in Fig.~\ref{fig:planning_comparison} we show one instance of global path planning for the mall map (Fig.~\ref{subfig:map_mall}). In this figure, it is apparent that our method is more efficient in finding a path due to the reduction of the search area depicted in green.
%We benchmark with the three path planners regarding the planning time, path length, and path smoothness on the map appearing in Fig.~\ref{fig:maps_bench}. The results appear in Table~\ref{tab:sim_run_time_comparision},~\ref{table-5-2}, and~\ref{table-5-8}, accordingly.

\subsection{Experiments on Real Quadrupeds}\label{Sec:realexp}
\begin{figure}[ht!]
    \centering
    \includegraphics[width=.3\textwidth]{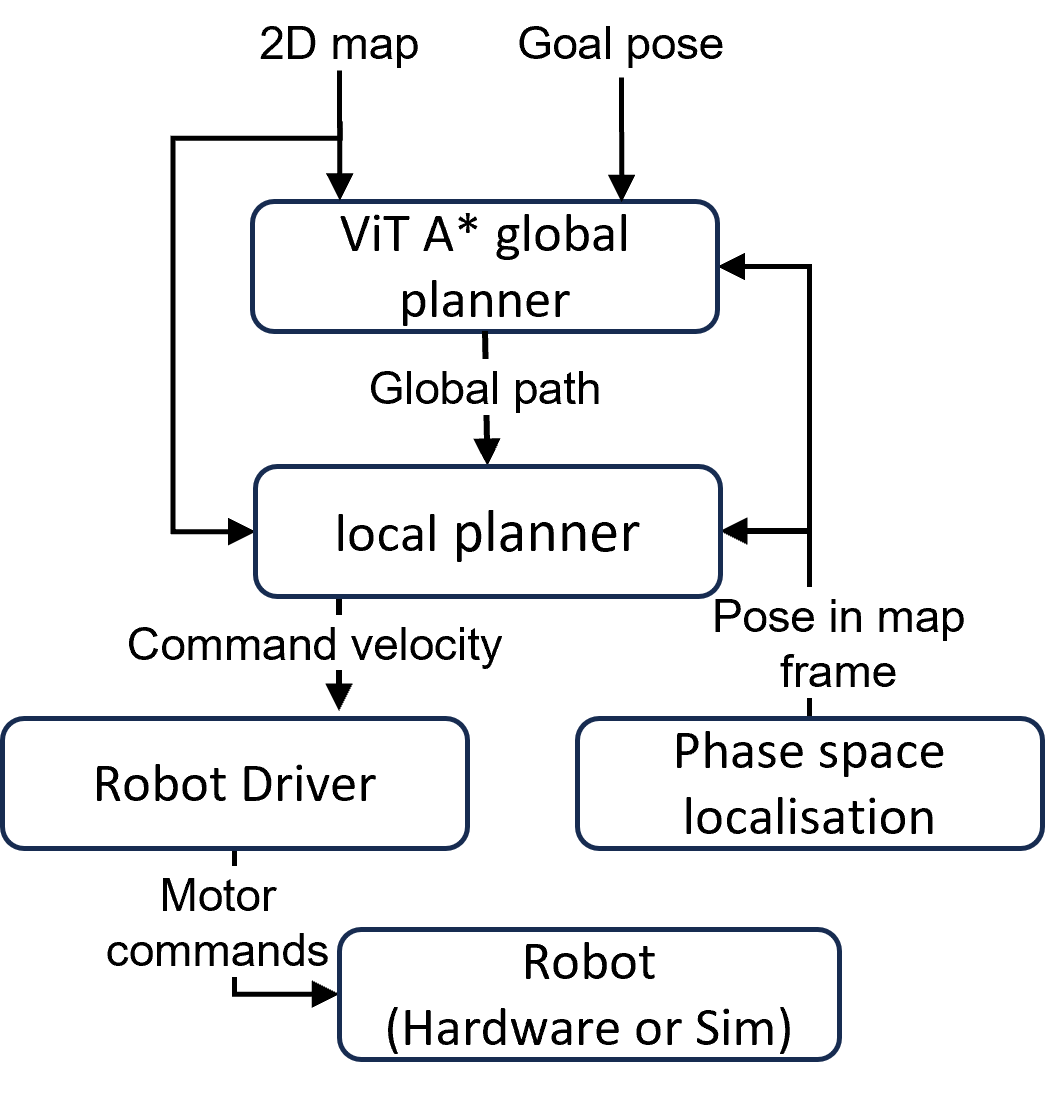}
    \caption{The schematic structure of the navigation stack for the  real robot.}
    \label{fig:planner_arch}
\end{figure}

To integrate the ViT-based $A^*$ path planner with the legged robot's navigation system, the planner is incorporated into an existing 2D ROS Navigation stack\footnote{http://wiki.ros.org/navigation} as a global path planner module. The overall architecture of the navigation stack is illustrated in Fig.~\ref{fig:planner_arch}. 

Within the stack, the  ViT-$A^*$ module generates the globally optimal path given the occupancy map. This path is then refined via the local planner to ensure compliance with the robot's kinodynamic constraints. In this case, the Timed-Elastic-Band (TEB)\cite{rosmann2012trajectory, rosmann2013efficient} local planner is employed. To mitigate the impact of state estimation on the quality of evaluation of the path planning module, an external tracking system, specifically the Phasespace tracking cameras\footnote{https://www.phasespace.com/}, is utilized. These cameras offer real-time localization for the robot at 960 Hz. Examples of the robots with active LED tracking markers can be seen in Fig.~\ref{fig:robots_with_PS_markers}. 

Note that in order to integrate the Neural $A^*$ module as a global planner plugin, several modifications were necessary to interface it with the rest of the navigation stack. Firstly, the navigation stack utilizes the \textit{OccupancyGrid} message to encode the map, which represents each cell in the map with the obstacle probability $p_{o}\in[0,100]$. However, the current version of the  ViT-$A^*$ module can only take a binary occupancy map as input, where each $\text{cell}_i\in \{0, 1\}$. To convert the map, an occupancy threshold $t$ is applied, in accordance to 
\begin{equation}
    \text{cell}_i = \begin{cases}
			1, & p_{o} \geq t\\
                0, & \text{otherwise}
		 \end{cases}
   \label{eq:map_convert}
\end{equation}
Furthermore, the ViT-$A^*$ module only produces paths as a sequence of positions, without considering the robot's orientation. Consequently, to generate the 3 Degrees of Freedom path compatible with the ROS stack, a simple forward-only orientation filter is implemented. This filter defines the orientation as the direction facing forward along the path, given a sequence of positions $\{v_i\}$ extracted from a global path, excluding the start and goal positions $v_s$ and $v_g$ (as their orientations are fixed by inputs - i.e. the current pose of the robot and desired goal pose).
Hence, the orientation $\theta_i$ is defined as:
\begin{equation}
    \theta_i = cos^{-1} \frac{v_i \cdot v_{i + 1}}{|v_{i}| |v_{i + 1}|}
    \label{eq:forward_oritation_filter}
\end{equation}
To validate the complete pipeline, tests were conducted on two real robots: Boston Dynamics Spot and Unitree Go1. In these tests, a predefined map is fed into the ViT-$A^*$ module, which generates a global path. Subsequently, the robot executes the planned path via the ROS stack described in Figure~\ref{fig:planner_arch}. The effectiveness of the pipeline has been demonstrated in navigation scenarios around obstacles in our laboratory, as shown in Figure~\ref{fig:Go1_nav_around_obstacle}.

\section{CONCLUSION}\label{sec:conclusion}
In this study, we present the ViT-$A^*$ planning strategy that enables quadrupedal robots to autonomously and safely navigate in various and complex scenarios. Our proposed method builds upon recent advancements in differential planning and introduces a pre-processing model based on ViT, enabling our approach to handle maps of any size. The effectiveness of the proposed approach has been validated through successful comparison in simulation and on real quadrupedal robots across different scenarios. In future work, we intend to evaluate the performance of our method in outdoor or more complex settings (e.g., autonomous task planning~\cite{zhou2022teleman}) and explore the benefits of planning directly on an RGB map with a ground truth path designed by humans.

% \section*{Acknowledgments}

\addtolength{\textheight}{0cm}   % This command serves to balance the column lengths

%%%%%%%%%%%%%%%%%%%%%%%%%%%%%%%%%%%%%%%%%%%%%%%%%%%%%%%%%%%%%%%%%%%%%%%%%%%%%%%%
\bibliographystyle{IEEEtran}
\bibliography{IEEEabrv, humanoids_2023_jianwei-eric}

\end{document}